\documentclass[journal]{IEEEtranTIE}
\usepackage{graphicx}
\usepackage{cite}
\usepackage{picinpar}
\usepackage{amsmath}
\usepackage{url}
\usepackage{flushend}
\usepackage[latin1]{inputenc}
\usepackage{colortbl}
\usepackage{soul}
\usepackage{multirow}
\usepackage{array}
\usepackage{pifont}
\usepackage{color}
\usepackage{alltt}
\usepackage{makecell}
\usepackage[hidelinks]{hyperref}
\usepackage{enumerate}
\usepackage{siunitx}
\usepackage{breakurl}
\usepackage{epstopdf}
\usepackage{pbox}
\usepackage{amssymb}
\usepackage{subfigure}
\usepackage{pifont}
\usepackage{booktabs}
\usepackage[flushleft]{threeparttable}

\begin{document}
\title{Spatial-wise Dynamic Distillation for MLP-like Efficient Visual Fault Detection of Freight Trains}

\author{
	\vskip 0.5em
	
	Yang~Zhang, Huilin~Pan, Mingying~Li, An~Wang, Yang~Zhou, and Hongliang Ren

\thanks{
	This work was supported in part by the State Key Laboratory of Novel Software Technology (Grant KFKT2022B38), Hubei Key Laboratory of Modern Manufacturing Quality Engineering (Grant KFJJ-2022014) and the Ph.D. early development program of Hubei University of Technology (Grant XJ2021003801). (Corresponding author: Huilin~Pan).
	
	Y. Zhang, H. Pan, M. Li, and Y. Zhou are with the School of Mechanical Engineering, 
	Hubei University of Technology, Wuhan 430068, China (e-mail: yzhangcst@hbut.edu.cn; hlp@hbut.edu.cn; 102200036@hbut.edu.cn; z-g99@hbut.edu.cn). 
	
	H. Pan, A. Wang and H. Ren are with the Department of Electronic Engineering, The Chinese University of Hong Kong, Hong Kong, China (e-mail: huilinpan@cuhk.edu.hk; wa09@link.cuhk.edu.hk; hlren@ieee.org).	

    Y. Zhang is also with the National Key Laboratory for Novel Software Technology, Nanjing University, Nanjing 210023, China.	(e-mail: yzhangcst@smail.nju.edu.cn)
	}
}

\maketitle

\begin{abstract}
Despite the successful application of convolutional neural networks (CNNs) in object detection tasks, their efficiency in detecting faults from freight train images remains inadequate for implementation in real-world engineering scenarios. 
Existing modeling shortcomings of spatial invariance and pooling layers in conventional CNNs often ignore the neglect of crucial global information, resulting in error localization for fault objection tasks of freight trains.
To solve these problems, we design a spatial-wise dynamic distillation framework based on multi-layer perceptron (MLP) for visual fault detection of freight trains. We initially present the axial shift strategy, which allows the MLP-like architecture to overcome the challenge of spatial invariance and effectively incorporate both local and global cues. We propose a dynamic distillation method without a pre-training teacher, including a dynamic teacher mechanism that can effectively eliminate the semantic discrepancy with the student model. Such an approach mines more abundant details from lower-level feature appearances and higher-level label semantics as the extra supervision signal, which utilizes efficient instance embedding to model the global spatial and semantic information. In addition, the proposed dynamic teacher can jointly train with students to further enhance the distillation efficiency. Extensive experiments executed on six typical fault datasets reveal that our approach outperforms the current state-of-the-art detectors and achieves the highest accuracy with real-time detection at a lower computational cost.
\end{abstract}

  
\markboth{IEEE TRANSACTIONS ON INDUSTRIAL ELECTRONICS}%
{}

\definecolor{limegreen}{rgb}{0.2, 0.8, 0.2}
\definecolor{forestgreen}{rgb}{0.13, 0.55, 0.13}
\definecolor{greenhtml}{rgb}{0.0, 0.5, 0.0}

\section{Introduction} 

\begin{figure}[!t] \centering   
\subfigure[] {
\label{fig1:a}     
\includegraphics[width=0.23\columnwidth]{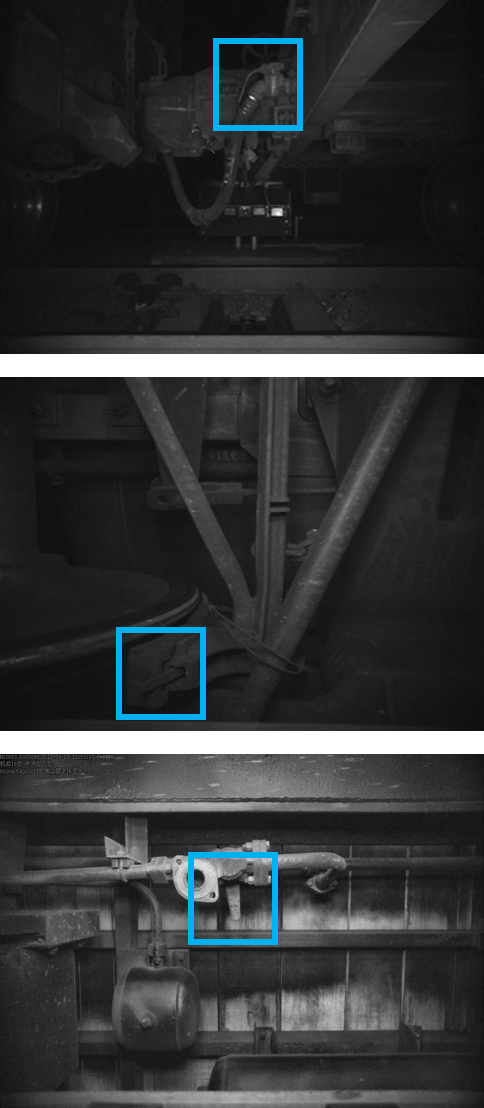}  
} \hspace{-1.2em}
\subfigure[] {
 \label{fig1:b}     
\includegraphics[width=0.23\columnwidth]{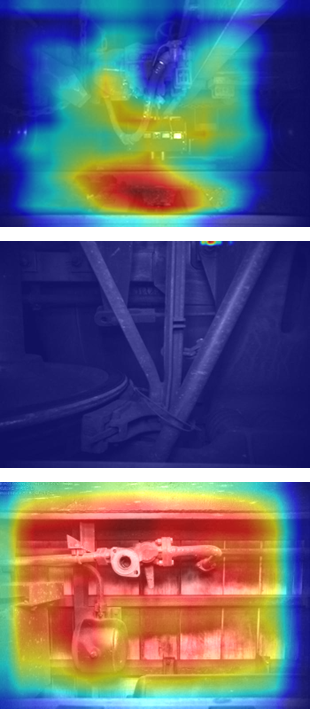}  
} \hspace{-1.2em}  
\subfigure[] { 
\label{fig1:c}     
\includegraphics[width=0.23\columnwidth]{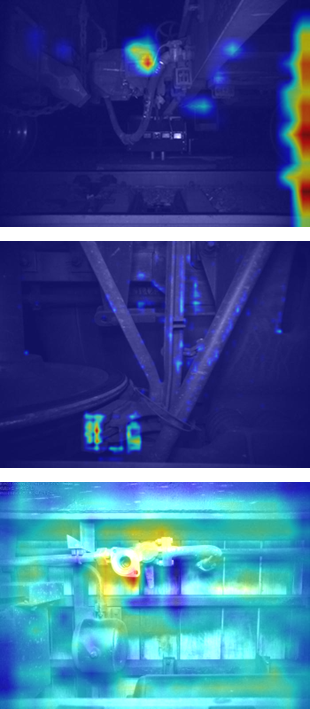}     
} \hspace{-1.2em}   
\subfigure[] { 
\label{fig1:d}     
\includegraphics[width=0.23\columnwidth]{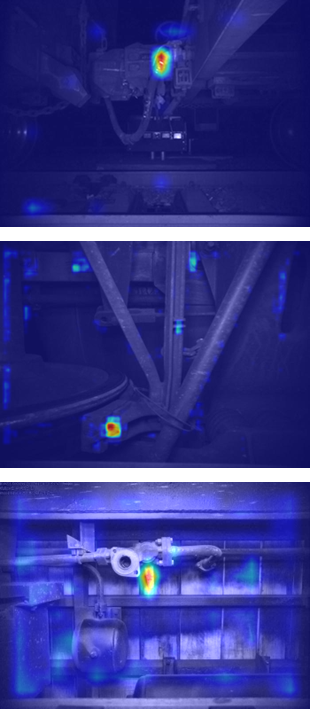}     
}   
\caption{Visualization results of activation maps from various detectors. (a) Ground truth annotations (blue) for each fault image. (b) CNN-based teacher-student distillation detector. (c) CNN-based self-distillation detector. (d) Our proposed MLP-like self dynamic distillation detector. The intensity of the feature response increases from blue to red. As the activation maps exemplify, our method can more accurately capture the region of interest due to the efficient encoding mechanism.}     
\label{fig1: activation map}
\end{figure}

\IEEEPARstart{V}{isual} fault detection of the train braking system is a crucial task to ensure the safe operation of the railway. 
In the domain of visual fault detection, machine learning-based detectors~\cite{Sun.J, Liu_tim} are gradually supplanting manual detection, which can eliminate human subjectivity while accomplishing greater accuracy and efficiency. As a branch of machine learning, deep learning techniques, especially convolutional neural networks (CNNs), have significantly boosted the performance of image recognition tasks. The CNNs typically emphasize local features through convolutional operations and pooling, which may disregard the broader context and correlation between faulty parts and the background. As shown in Fig.~\ref{fig1:a}, the intricate distributions and volume variations among fault elements and train bodies face challenges for CNN-based methods, specifically due to inherent spatial invariance, resulting in less accurate identification of fault regions. Therefore, capturing global cues and modeling spatial cues are the most critical issues for visual fault detection of freight trains.

Recently, multi-layer perceptrons (MLP) based methods and residual connections have emerged as a new trend in visual recognition tasks. Without using CNNs and self-attention mechanism, MLP-mixer~\cite{MLP-Mixer} uses matrix transpose and self-attention mechanism to demonstrate that pure MLP architecture can also achieve competitive results in image classification. Meanwhile, ResMLP~\cite{ResMLP} simplifies the token mixture MLP into a single fully connected layer and utilizes residual connections to create a more profound architecture to improve detection accuracy. These MLP-based methods address the issue of spatial invariance in CNNs, which are also effective at handling the local features of images. However, the high computational cost makes these methods challenging to deploy in field scenarios with limited computing resources.

As an emerging approach for obtaining an efficient framework, knowledge distillation shows excellent promise in practical scenarios with limited resources and power. 
Conventional knowledge distillation~\cite{Lad, LD_2022, ReviewKD, CWD, MGD} involves training teacher and student networks, where the purpose is to raise the capability of the student by delivering knowledge from the complex teacher network. Nevertheless, this method heavily relies on the effectiveness of the ideal teacher network, which may hardly be satisfied in real-world scenarios, and training the teacher model alone can considerably reduce the training efficiency. Alternative teacher-free techniques have been proposed (\textit{e.g.}, self-distillation~\cite{LGD}, label regularization~\cite{LabelEnc}) to circumvent these challenges and minimize the dependence on pre-trained teachers. In Fig.~\ref{fig1:b}, the distillation effect can be impacted by the inability of the CNN-based teacher-student distillation detector to achieve complete adaptation between them. In Fig.~\ref{fig1:c}, the CNN-based self-distillation detector cannot guarantee the effect of the distillation process due to spatial invariance modeling flaws of CNNs. Although these methods have demonstrated specific improvements, the detection efficiency and performance of this collaborative training approach still fall short of meeting the demands of fault detection in freight train scenarios.

To alleviate these restrictions, we propose an innovative \textbf{s}patial-wise \textbf{d}ynamic \textbf{d}istillation method for \textbf{f}ault \textbf{d}etection of \textbf{f}reight \textbf{t}rain \textbf{i}mages (SDD FTI-FDet). We present a lightweight MLP-based backbone with an axial shift strategy that can effectively mitigate the spatial invariance limitation of CNNs while capturing a greater extent of global and local spatial dependencies. Likewise, we reformulate the knowledge transfer process of spatially global information in the dynamic teacher without depending on intricate CNNs. The MLP-like framework and efficient instance embedding technique can augment the ability of the student detector to understand global spatial and semantic information to a certain extent. Furthermore, we design a feature adaptor to remap the encoded interaction embeddings into 2D space as final instructive knowledge. The introduction of the dynamic teacher enables joint training of teacher and student that bridges the semantic discrepancy. Our method improves detection efficiency and relieves the restriction of pre-training teachers in resource-constrained scenarios. 
Notably, on six typical fault datasets, our SDD FTI-FDet achieves the highest accuracy with real-time detection compared with state-of-the-art detectors while significantly reducing computational expense and maintaining the almost smallest model size of 74.5MB.

The main contributions are summarized as follows:
\begin{enumerate}[1)]
    \item We explore eliminating the limitation of spatial invariance of CNNs and utilizing a dynamic teacher to transfer spatial and semantic cues to improve performance.
    \item We apply a novel lightweight MLP-based backbone for resource limitations, and the introduction of the axial shift strategy can prevent the loss of global information.
    \item We design an optimized dynamic distillation without a pre-training teacher through efficient instance embedding and jointly train with students to boost detection efficiency.
    \item Extensive quantitative and qualitative analyses demonstrate that our method performs best and enables real-time fault detection with the lowest computational cost for freight train images.
\end{enumerate}


\section{Related works}\label{sec: related_works}
\subsection{Fault Detection for Freight Train Images}
In recent years, the application of computer vision technology in visual fault detection of freight trains has greatly improved detection efficiency. These detectors can be divided into two categories: traditional and deep learning-based methods. The formers primarily design features manually and use image processing methods combined with machine learning to realize fault detection~\cite{Sun.J, Liu_tim, Sun}. 
For instance, Liu et al.~\cite{Liu_tim} integrated the cascaded detection with the gradient-coded co-occurrence matrix features to inspect the bogie block key. 
These traditional methods are not universally applicable, and detection algorithms often must be redesigned for different scenarios. 
In contrast, deep learning methods have proven to be highly effective in extracting extensive and abstract features, enabling them to tackle intricate and demanding problems within the realm of machine vision. 
Chen et al.~\cite{Chen_tcyb} presented a novel end-to-end deep learning framework to fuse features and structural information in a weighted and recursive manner to achieve train component detection. 
Zhang et al.~\cite{Zhang_acess} used the mosaic data augmentation and the \textit{k}-means clustering to increase the data samples and generate high-quality prior bounding boxes. The above methods address the issues of multi-category fault detection and low detection accuracy, but the high computational cost makes it challenging for deployment in scenarios with constrained resources.

\subsection{General Object Detection}
\subsubsection{CNN-based}
As the mainstream method in the current deep neural network structure, CNN performs well in image classification, object detection, and segmentation tasks. The CNN detectors can be broadly classified into two categories: two-stage methods and one-stage detectors. Two-stage detectors such as Faster RCNN~\cite{Faster_R-CNN}, Libra R-CNN \cite{Libra_R-cnn}, and Sparse R-CNN~\cite{Sparse_R-CNN} first employ a region proposal network to effectively differentiate between foreground and background regions. Subsequently, the candidate regions are precisely localized and regressed in the second stage to improve accuracy. One-stage detectors~\cite{Centernet, FCOS, Focal_Loss, GFL, YOLOF, YOLOX} discard the candidate stage and directly generate candidate boxes and complete localization and classification. Compared with the two-stage methods, one-stage methods improve the detection efficiency by sacrificing part of the detection accuracy. In order to reduce computational expenses, certain techniques favor point-based representations over bounding boxes for attaining more accurate localization.

\begin{figure*}[!t]
	\centering
	\includegraphics[width=6.9in]{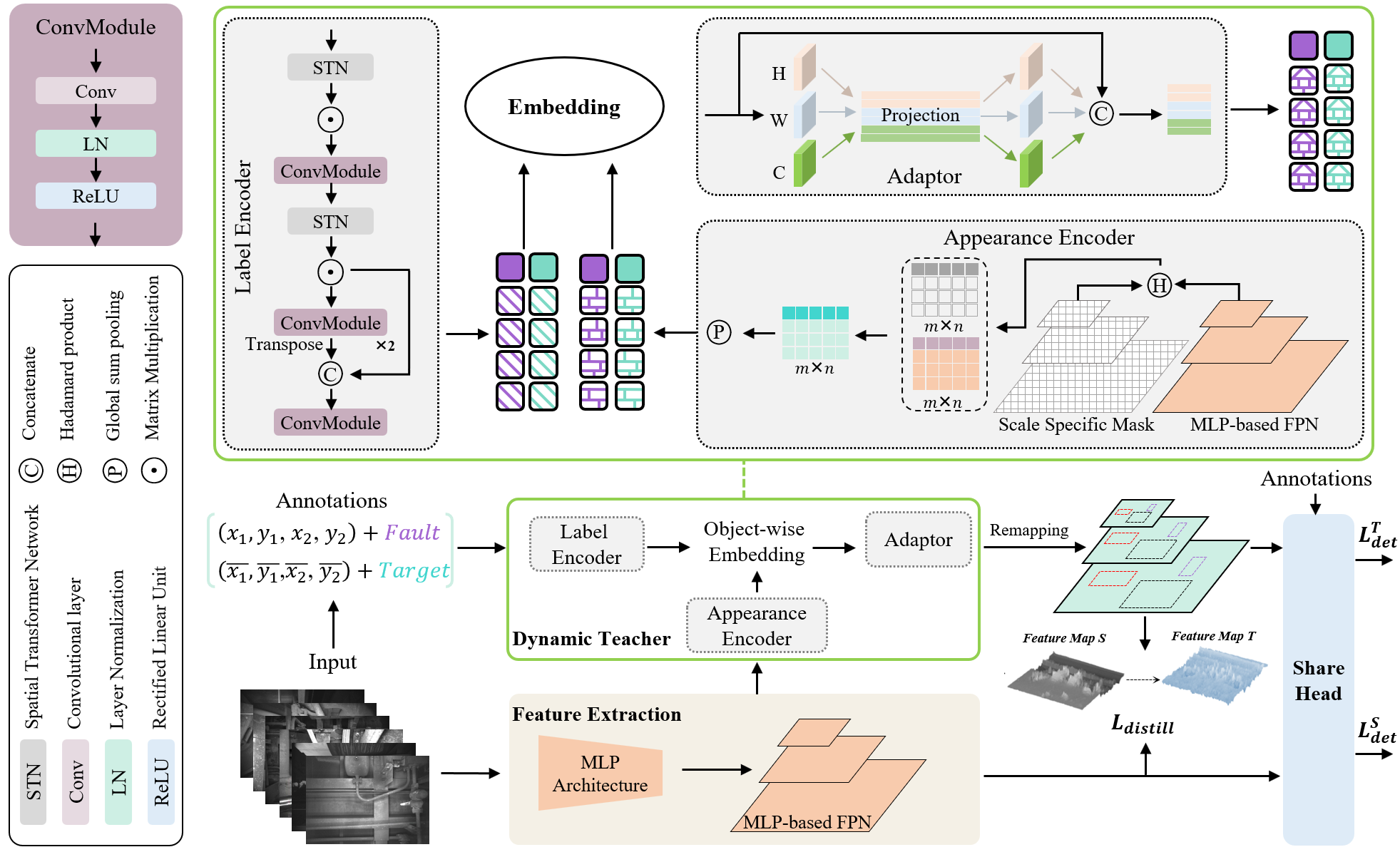}
	\caption{Overview of the proposed MLP-like spatial-wise dynamic distillation method. We adopt a novel dynamic teacher architecture comprising three modules: label encoder, appearance encoder, and feature adaptive interaction.  The dynamic teacher enables joint teacher-student training, which generates instructional representations from ground truth annotations and feature pyramids during the training stage. Meanwhile, the instructional representations and the student features were individually supervised by detection loss $L_{det}^{T}$ and $L_{det}^{S}$ at each feature scale to guarantee representational capability. We denote $(x_{1}, y_{1}, x_{2}, y_{2})$ as the ground-truth box location normalized by image size.}
	\label{fig2: framework}
\end{figure*}
\subsubsection{MLP-based}
Recently, models based on the transformer and MLP have shown great potential to solve contemporary problems in the image classification task. Detection transformer (DETR)~\cite{DETR,Deformable_DETR} introduces the transformer into the target detection task and combines it with CNN to directly predict the final detection result. This approach does not require any post-processing stage and achieves end-to-end object detection. MLP-Mixer~\cite{MLP-Mixer} achieves the integration of long-distance dependencies by employing matrix transposition and token-mixing projection to obtain a global receptive field. However, this method fails to fully utilize local information. To resolve this problem, Lian et al.~\cite{AS-MLP} proposed axial shifted MLP to use an axial shift strategy to combine location features in different spaces to obtain more local dependencies and improve performance. Compared with DETR-based architectures, AS-MLP~\cite{AS-MLP} surpasses all MLP-based methods and accomplishes excellent performance in downstream tasks. Nevertheless, these methods are designed for environments with sophisticated computing resources, which are not available or affordable in real-world scenarios. 

\subsection{Knowledge Distillation for Object Detection}
Knowledge distillation can learn a compact and efficient student model guided by an excellent teacher network, effectively reducing model size while maintaining model accuracy. Channel-wise distillation (CWD)~\cite{CWD} makes the distillation process more efficient by normalizing the activation maps of each channel and minimizes the Kullback-Leibler (KL) divergence between the teacher and student networks. Masked generative distillation (MGD)~\cite{MGD} creates a random mask and utilizes the characteristics of the students to generate more robust characteristics of the teacher to further enhance the representation ability of the students. Localization distillation (LD)~\cite{LD_2022} introduces valuable localization regions and transfers classification and localization knowledge separately to improve the distillation efficiency. The above methods are all proposed based on models of powerful teachers, which are often unavailable in real-world scenarios. 

Recently, teacher-free methods have been proposed to mine instructive knowledge from the model itself or hard labels. For example, Hao et al.~\cite{LabelEnc} regarded the ground-truth label as an intermediate layer representation with category and position information, which serves as a supervisory signal during training. Label-guided distillation (LGD)~\cite{LGD} employs efficient instance embeddings to design student performance and regular labels as implicit teachers and also achieves impressive performance on detection tasks. Nevertheless, these methods dismiss the significance of spatial location information carried by 2D feature representations. 

Unlike the aforementioned teacher-based distillation methods for CNN, we introduce a dynamic distillation method with an MLP-like structure and propose an efficient instance embedding method to further capture more spatial information while improving distillation efficiency and quality.

\section{Method}\label{sec: method}
In this section, the proposed SDD FTI-FDet alleviates the semantic discrepancy between teacher and student models by employing dynamic teacher-student joint training. We also introduce an efficient instance embedding coding method and a feature adaptor to facilitate knowledge transfer in two-dimensional space. The overall framework, depicted in Fig~\ref{fig2: framework}, comprises a lightweight MLP-based backbone for feature extraction in the student network and a dynamic teacher network consisting of a label encoder, appearance encoder, and interactive embedding encoder. We incorporate a dynamic distillation strategy that enforces high-order knowledge consistency between the outputs of the teacher and student networks. Notably, our method only utilizes the student detector during inference without additional computational costs.

\subsection{Lightweight MLP-based Backbone}\label{Lightweight Backbone}
The backbone is the most computationally expensive structure in the network, and its main function is feature extraction, which largely determines the final size of the model. A lightweight backbone is indispensable since fault detection is usually in an environment with limited field resources. Recent studies have shown that MLP can achieve strong performance at a small model size, and some MLP-based architectures~\cite{AS-MLP} have performed well on computer vision tasks to a level comparable to that of CNN-based. 
However, the characteristics of the fully connected layer lead to a large number of parameters and calculations in the network, which is unsuitable for the field environment with limited resources. So, we design a lightweight MLP-based backbone FaultMLP, to capture local dependencies by transforming the channel of the features to obtain information flows in different axes.

In Table~\ref{backbone architecture}, FaultMLP adopts the core MLP block~\cite{AS-MLP}, which is mainly composed of the LayerNorm (LN) layer, axial shift operation, MLP, and residual connection. The input $X$ is first divided into $s$ splits in the horizontal and vertical direction. After the axial shift operation, the output $Y_{i,j}$ is:
\begin{equation}\label{mlp}
    Y_{i,j}\!=\!\sum_{c = 0}^{C}\!w_c^h\!X_{i+\lfloor \frac{c}{\lceil C/s \rceil} \rfloor\!-\!\lfloor{\frac{s}{2}\rfloor}\cdot d,j,c}\!+\!\sum_{c = 0}^{C}\!w_c^v\!X_{i,j+\lfloor {\frac{c}{\lceil C/s \rceil}} \rfloor\!-\!\lfloor{\frac{s}{2}\rfloor}\cdot d,c},
\end{equation}
where $d$ is dilation rate. The $w^h$, $w^v \in \mathbb{R}^C$ are the learnable weights of channel projection in the horizontal and vertical directions, respectively. These are aggregated together to promote the information flow from various channels in the spatial dimension. When the shift size is 3, the input feature is divided into three parts and shifted by \{-1, 0, 1\} units along the horizontal direction, respectively. The features in the dashed box will then be removed and utilized for the following channel projection. In the vertical shift, the same operation is also performed. Information from several spatial locations can completely flow and interact in the next channel projection.

Given a fault image $I \in \mathbb{R}^{3 \times H \times W}$, first performs a partition operation to divide the image into multiple 4$\times$4 patches and then pass through linear embedding and MLP blocks to obtain the output features of $C \times \frac{H}{4} \times \frac{W}{4}$. The FaultMLP performs group-shift-merge operations on features in both horizontal and vertical directions to enhance the information exchange of the network in both directions. Finally, the projection operation maps the features into a linear layer to fully communicate information from different spatial locations. 

\begin{table}[!t]
\centering
\small
\caption{The detailed configurations of FaultMLP. the size of the input image is 224$\times$224. Concat n$\times$n: concatenation of n$\times$n neighboring features in a patch. Shift size (5, 5): shift size in the horizontal and vertical directions is 5.}
\renewcommand{\arraystretch}{1.1}
\setlength{\tabcolsep}{2mm}{
\begin{tabular}{ccc}
\toprule
Stage & \begin{tabular}[c]{@{}c@{}}downsp. rate \\ (output size)\end{tabular} & Modules  \\
\midrule
\multirow{3}{*}{1} & \multirow{3}{*}{\begin{tabular}[c]{@{}c@{}}4$\times$\\ (56$\times$56)\end{tabular}} & Concat 4$\times$4, 64-d, LN  \\
& & $\begin{bmatrix}\text{Shift size (5, 5),}\\\text{dim 64}\end{bmatrix}$ $\times$ 2  \\
\midrule
\multirow{3}{*}{2}  & \multirow{3}{*}{\begin{tabular}[c]{@{}c@{}}8$\times$\\ (28$\times$28)\end{tabular}} & Concat 2$\times$2, 128-d , LN \\
& & $\begin{bmatrix}\text{Shift size (5, 5),}\\\text{dim 128}\end{bmatrix}$ $\times$ 2   \\
\midrule
\multirow{3}{*}{3}  & \multirow{3}{*}{\begin{tabular}[c]{@{}c@{}}16$\times$\\ (14$\times$14)\end{tabular}}  & Concat 2$\times$2, 256-d , LN \\
& & $\begin{bmatrix}\text{Shift size (5, 5),}\\\text{dim 256, head 12}\end{bmatrix}$ $\times$ 2 \\
\midrule
\multirow{3}{*}{4} & \multirow{3}{*}{\begin{tabular}[c]{@{}c@{}}32$\times$\\ (7$\times$7)\end{tabular}}  & Concat 2$\times$2, 512-d , LN \\
& & $\begin{bmatrix}\text{Shift size (5, 5),}\\\text{dim 512, head 24}\end{bmatrix}$ $\times$ 2 \\
\bottomrule
\end{tabular}}\label{backbone architecture}

\end{table}

\subsection{Dynamic Teacher}\label{Dynamic Teacher}
\subsubsection{Label Encoder}\label{Label Encoder}
The label encoder is designed to encode the ground-truth labels into latent embeddings, which can relieve the knowledge discrepancy between teacher and student and assistant supervision of student detectors during training. In our approach (depicted in Fig~\ref{fig2: framework}), we first concatenate the normalized ground-truth box coordinates $\left(x_1, y_1, x_2, y_2\right)$ and category labels to form a label descriptor ${l}_i$, where $i$ represents the context index of $N$ objects in the image. We then define $\mathcal{L}=\left\{{l}_i \in \mathbb{R}^C\right\}_{i=1}^N$, $(1 \leq i \leq N)$ as the set of descriptions of all objects in the image, where $C$ represents the number of channels. It is worth noting that when the label descriptor set of each batch undergoes a certain spatial transformation, the semantic information remains unchanged. 
Specifically, we initialize $\mathcal{L}$ and then employ spatial transformer networks (STN) for spatial transformation at the input and feature levels. Meanwhile, features before and after transformation are aligned by matrix multiplication, which enables the modeling of local-global relationships. Furthermore,  we replace BatchNorm with LN to adapt to the constraints of small-batch detection during training.

\subsubsection{Appearance Encoder}\label{Appearance Encoder}
Considering the semantic discrepancies of heterogeneous inputs (images vs. labels), we extract the necessary appearance embeddings from the feature pyramid of a student detector that includes intra-object relations. We set mask pooling at corresponding sizes based on the feature pyramid output: $\mathcal{M}_p= {\left\{\mathrm{m}_i\right\}_{i=1}^N} \in \mathbb{R}^{{H}_p \times {W}_p}$, where $\mathcal{M}_p$ represents mask pooling at $p$-th scale, and each mask pool contains $N$ objects and one virtual object (covering the entire image). At each of the pyramid levels, object-wise masks at the input level are down-scaled to correspond with resolution to become scale-specific ones. For each object, appearance embeddings are computed from the Hadamard product between the feature matrix and object masks. The feature matrix can be obtained by computing the projected feature map $X_p \in \mathbb{R}^{{H}_p \times {W}_p \times C}$. Therefore, appearance embedding $\mathcal{A}_p\in \mathbb{R}^{C}$ at the $p$-th is defined as:
\begin{equation}
\label{Eq1}
\setlength\abovedisplayskip{7pt} 
\setlength\belowdisplayskip{7pt}
    \mathcal{A}_p = \mathcal{F}_{\text {proj }}\left(X_p\right) \ast \mathcal{M}_p,
\end{equation}
where $\ast$ represents Hadamard product, and $\mathcal{F}_{\text {proj}}$ is a single $1\times1$ convolutional layer. We perform the same operation for each object at $p$-th scale to extract the appearance embedding of the image at the corresponding resolution. Then, the taken scale appearance embeddings will be processed by global sum pooling for normalization to achieve object size-invariant.

\begin{figure} 
\centering    
\subfigure[] {  
\label{fig3:a}  
\includegraphics[width=0.31\columnwidth]{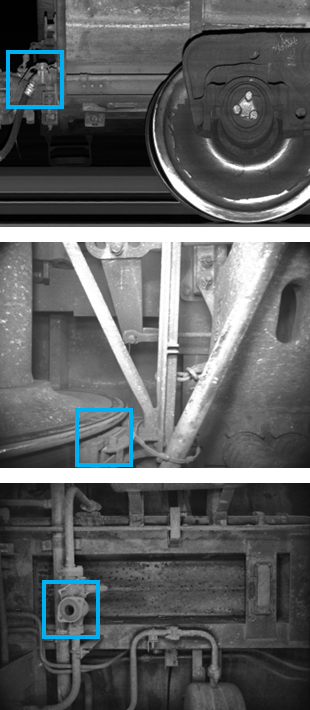}  
} \hspace{-1.1em}
\subfigure[] { 
\label{fig3:b}  
\includegraphics[width=0.31\columnwidth]{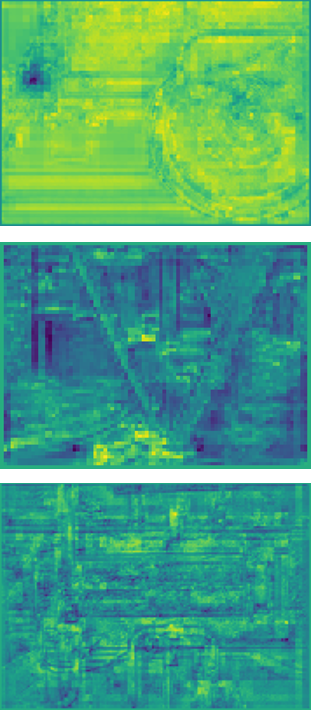}  
} \hspace{-1.1em}  
\subfigure[] {  
\label{fig3:c}  
\includegraphics[width=0.31\columnwidth]{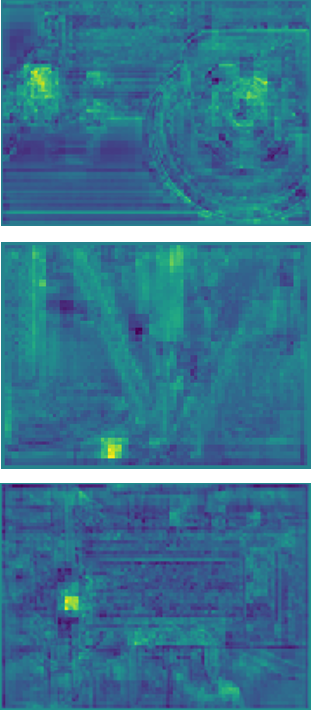}     
} 
\caption{Visualization comparison results of feature maps after student feature pyramid network on three typical fault datasets. (a) Ground truth annotations (blue box) for each fault image. (b) The feature maps with the original embedding method.  (c) The feature maps with the proposed feature adaptor. The fault area in (a) is reliably detected with the proposed novel encoding method. }     
\label{fig3: feature map}     
\end{figure}

\subsubsection{Interactive Embedding Encoder}\label{Interactive Embedding Encoder}
It is worth noting that after we obtain the label embedding and appearance embedding, another problem we need to consider is how to realize the feature adaptive interaction of different forms inputs.
A natural thought is to query by key and value and establish the adaptive interaction between label embedding and appearance embedding through information reorganization. We introduce priors: appearance embeddings as queries $Q$ and scale-invariant label embeddings as keys $K$ and values $V$. Before information recombination, $Q$, $K$, and $V$ are transformed through a linear layer $f_{l}$, and then we apply a multi-head self-attention block~\cite{MHSA} to process the appearance embedding $a_{i}$ and the label embedding $l_{j}$ to obtain the intermediate output. The intermediate output of the $i$-th object is denoted as:
\begin{equation}
\label{Eq1}
    u_{i} = {\sum_{j = 0}^{n}{{\left\{ \delta\left( \frac{{f_{l}Q\left( a_{i} \right) \cdot f}_{l}{K\left( l_{j} \right)}}{\sqrt{C}} \right) \right\} } f_{l}V\left( l_{j} \right)}},
\end{equation}
where ${\sqrt{C}}$ denotes dot-product attention scaling factor and $\delta$ represents the $SoftMax$ function. The interaction embedding $\mathbf{E}_p$ can be obtained by concatenating the intermediate outputs generated by all heads and performing linear projection $f_{p}$: 
\begin{equation}
\label{Eq1}
    \mathbf{E}_p=f_{p}\left(\left[u_{i}^1 ; u_{i}^2 ; \ldots \right]\right),
\end{equation}
where $[;]$ represents a concatenation operation along the channel dimension. 
As mentioned before, we iterate over feature layers at all scales to capture abundant interaction embeddings.


\subsection{Dynamic Distillation Strategy}
\label{Dynamic Distillation}
\subsubsection{Feature Adaptor}\label{Feature Adaptor}
As mentioned above, our proposed dynamic teacher extracts interactive embeddings from the output layer of the label and student pyramids as guiding knowledge, which solves the problem of impaired distillation performance owed to the knowledge discrepancy between the teacher and student models. 
Another critical problem is how to map the 1D interaction embedding to the 2D feature space to achieve spatial alignment and fit the student representation to the activation neurons of the instructional knowledge. 

Considering that traditional MLP-based methods suffer from a limitation where they encode spatial information along flattened dimensions, which results in the loss of positional information stored in 2D features. In Fig~\ref{fig3:b}, this loss of information poses challenges in ensuring the accuracy of subsequent operations.
Encoding adaptive interaction embeddings directly in the spatial dimension is a more effective alternative. This dynamic adaptation enables the teacher network to continuously refine its knowledge based on the student network's evolving capabilities to improve the student network's performance. Following the overall MLP-like architecture, inspired by the strategy of ~\cite{VIP}, we design a novel feature adaptor to model the spatial information between interaction embeddings in the dynamic teacher to address this issue. This alternative approach directly encodes adaptive interaction embeddings in the spatial dimension, which proves to be a more effective solution in Fig~\ref{fig3:c}.

Given the inputs $\mathbf{E}_p \in \mathbb{R}^{H \times W \times C}$ at $p$-th scale, we first separate $\mathbf{E}_p$ into $s$ dimension along the channel and conduct $(H, W, C)\rightarrow(C, W, H)$ operation on each dimension to generate height-channel permutation. We concatenate them along the channel dimension as the output of the permutation operation and use fully connected layers to mix height information. Meanwhile, the height replacement operation is performed again to restore the original dimension to obtain the output $\mathbf{X}_{H}$. Similarly, the same operation for the width to get $\mathbf{X}_{W}$. For the channel dimension, the encoding in the channel direction can be linearly projected to obtain  $\mathbf{X}_{C}$ by introducing a fully connected layer. Finally, we connect the sum of the encoding information in three-dimensional $\{C, H, W\}$ by weighted aggregation method (following the setting in~\cite{VIP}) and then mix it through a fully connected layer to obtain the final output: 
\begin{equation}
\label{Eq1}
    \hat{\mathbf{X}}_{T}=\operatorname{FC}\left(\mathbf{X}_H+\mathbf{X}_W+\mathbf{X}_C\right).
\end{equation}
Specifically, for the student model, we adopt an adaptive head $f_{adapt}$ to perform feature adaptation on the intermediate layer output $F_{p}^{S}$ of the student detector, which can be defined as:
\begin{equation}
\label{Eq1}
     \hat{\mathbf{X}}_{S} = f_{adapt}\left( F_{p}^{S} \right).
\end{equation}
\subsubsection{Loss Function}\label{Loss Function}
For knowledge transfer between the obtained instructional representations and student features, we introduce logit distillation to enforce the student network outputs more similar predictions with the teacher network. The output distribution difference between the dynamic instructor and the adaptive student is calculated by applying KL during the training progress. The distillation loss $L_{distill}$ between the two probabilities is attained by:
\begin{equation}
\label{Eq1}
    L_{distill}=\frac{1}{N} \sum_{i=1}^N \mathrm{KL}\left(S(\hat{\mathbf{X}}_{T}, \tau), S(\hat{\mathbf{X}}_{S}, \tau)\right),
\end{equation}
where $\hat{\mathbf{X}}_{T}$ and $\hat{\mathbf{X}}_{S}$ is transforms into probability distributions utilzing $SoftMax$ function $S(\cdot, \tau) = SoftMax(\cdot/\tau)$ and $\tau$ is distillation temperature. 

Additionally, we further ensure the consistency between the guiding features obtained by the dynamic teacher and the student features by sharing the detection head.
In summary, the total training objective is:
\begin{equation}
\label{Eq1}
     L_{total}= L_{det}^{S}+L_{det}^{T} + \lambda L_{distill},
\end{equation}
where $\lambda=1$ is a trade-off for the distillation term throughout all experiments. It is notable that our SDD FTI-FDet can be effectively employed across a wide spectrum of detectors, transcending the limitations imposed by traditional categorizations such as one-stage and two-stage detectors.

\section{Experiments}\label{sec: experiments}

\begin{table*}[!t]
    \renewcommand{\arraystretch}{1.05}
    \caption{Comparison results with some state-of-the-art methods on six visual fault datasets from freight trains. The performance of the knowledge distillation methods is from the student model. To obtain the best performance from each detector, we fine-tune relevant parameters. \textbf{T}: Teacher. \textbf{S}: Student}
    \centering
    \small
    \setlength{\tabcolsep}{2mm}
    \begin{threeparttable}{
        \begin{tabular}{lcccccccccc}
        \toprule
        \multirow{2}{*}{{Methods}}& 
        \multirow{2}{*}{{Backbones}}& 
        \multirow{2}{*}{{$mAP$}}& 
        \multirow{2}{*}{{$AP_{50}$}}& 
        \multirow{2}{*}{{$AP_{75}$}}&
        \multirow{2}{*}{{$AR_{1}$}}&
        \multirow{2}{*}{{$AR_{10}$}}& 
        \multirow{2}{*}{\makecell[c]{Test Memory\\(MB)}}  &
        \multirow{2}{*}{\makecell[c]{Model\\size (MB)}}  &
        \multirow{2}{*}{\makecell[c]{Infer.\\(FPS)}}  \\
        & & & & & & &  &   \\
        \midrule
        \textbf{\textit{General object detector}} \\
        CenterNet \cite{Centernet}              &R18    &40.6   &66.5 &41.0 &52.0 &54.4  
            &1565 &\textbf{67.3}    &32.1\\ 
        Deformable DETR \cite{Deformable_DETR}  &R50    &32.6   &63.8 &30.2 &47.2 &51.0  
            &1727 &210.7   &14.0\\
        Faster RCNN \cite{Faster_R-CNN}		    &R50    &41.0   &66.1 &42.9 &52.3 &53.3   
            &1935 &167.7   &24.2 \\
        FCOS \cite{FCOS}		                &R50    &47.3   &73.8 &50.1 &55.7 &58.2   
            &1665 &131.2   &23.4\\
        GFL \cite{GFL}                          &R50    &46.5   &72.3 &48.3 &55.6 &59.0    
            &1705 &130.0   &25.2\\
        Libra R-CNN \cite{Libra_R-cnn}	        &R18    &39.4   &65.5 &40.8 &52.8 &55.6  
            &1889 &82.6    &27.1\\
        RetinaNet \cite{Focal_Loss}             &R18    &37.0   &63.8 &37.3 &49.9 &53.6   
            &1647 &81.5    &29.2\\
        RetinaNet \cite{Focal_Loss}             &SwinT-t &40.1   &67.2 &41.3 &52.5 &55.9  
            &1823 &151.0   &19.5\\
        Sparse R-CNN \cite{Sparse_R-CNN}	    &R50    &46.7   &70.5 &49.4 &55.2 &58.5 
            &1829 &269.0   &19.8\\
        VarifocalNet \cite{VarifocalNet}        &R50    &47.0   &72.6 &49.4 &56.3 &59.5 
            &1715 &133.2   &19.4\\
        YOLOF \cite{YOLOF}                      &R50    &47.2   &73.8 &50.9 &55.1 &58.6 
            &1701 &172.5   &31.0\\
        YOLOX \cite{YOLOX}                      &CSPDarkNet &44.4   &69.3 &46.3 &55.5 &57.3 
            &1649 &84.9    &29.6\\
        \midrule
        \textbf{\textit{T-S knowledge distillation}} \\
        LAD~\cite{Lad}    & T: R101\quad S: R50   & 38.1  & 64.2  & 40.3  &  55.1     & 58.6       
            & 1853      & 257.2   & 18.3  \\
        LD~\cite{LD_2022}    & T: R101\quad S: R18   & 49.5  & 88.8  & 46.1  &  58.4     & 61.9     
        & 1821      & 154.4   & 33.3  \\
        RKD~\cite{ReviewKD}   & T: R101\quad S: R18   & 31.3  & 47.9  & 35.5  & 34.5  & 34.9    
            & 1514  & 472.0   & 49.3  \\
        CWD~\cite{CWD}   & T: R101\quad S: R50   & 55.2  & 86.8  & 61.1  & 63.7  & 63.7   
            & 1447      & 399.3   & 37.1  \\
        MGD~\cite{MGD}   & T: R101\quad S: R50   & 54.3  & 86.8  & 59.2  & 63.8  & 63.8   
            & 1683  & 900.0   & 37.8  \\
        \midrule
        \textbf{\textit{Self knowledge distillation}} \\
        LabelEnc~\cite{LabelEnc} & T: \hspace{0.6em}--\hspace{0.6em} \quad S: R50   & 50.7  & 82.5  & 55.3  & 57.8  & 60.3      
            &  3267     & 388.8  & 35.3 \\
        LGD~\cite{LGD}   & T: \hspace{0.6em}--\hspace{0.6em} \quad S: R50      & 54.8  & 85.3  & \textbf{59.9}  & 60.9  & 62.5    
            & 1443  & 409.5  & \textbf{43.3}  \\
        \textbf{SDD FTI-FDet} (Ours)  & T: \hspace{0.6em}--\hspace{0.6em} \quad S: R50      & \textbf{55.2}  & \textbf{88.0}  & \textbf{58.8}  & \textbf{61.2}  & \textbf{62.8}    
            & \textbf{1149}  & \textbf{74.5}   & \textbf{38.8} \\
        \bottomrule
    \end{tabular}}
    \end{threeparttable}
 \label{sota}
\end{table*}

\subsection{Experiments Setup}
\subsubsection{Datasets}
\begin{figure}[!t]
	\centering
	\includegraphics[width=2.5in]{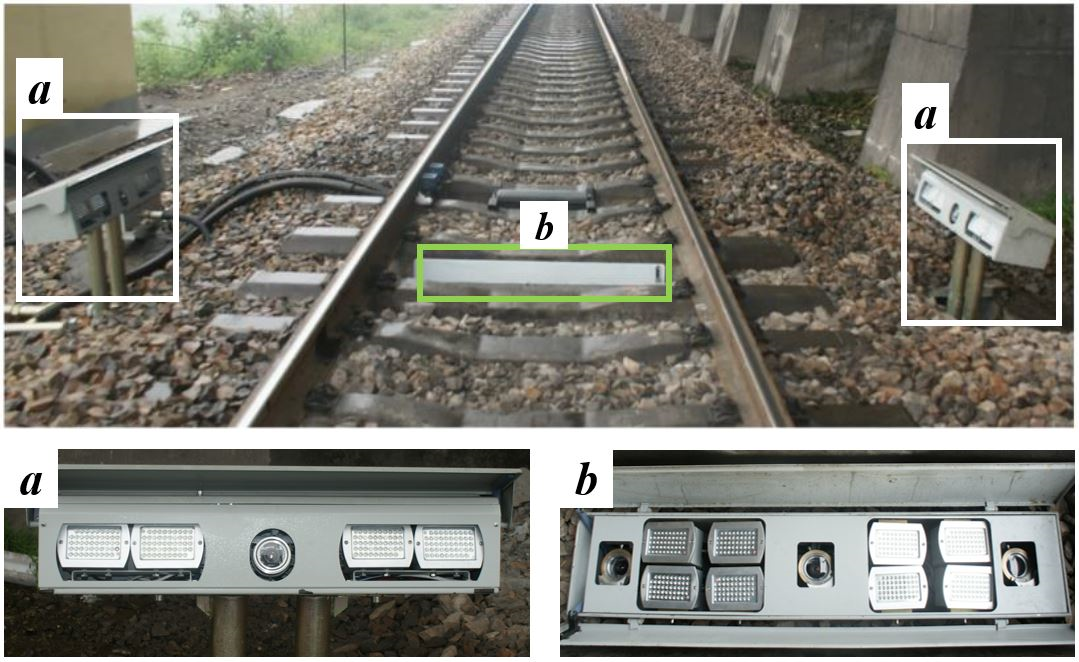}
	\caption{Hardware platform for fault detection. The image acquisition system consists of high-speed cameras and light arrays, which are integrated and placed between the trackside and the track to obtain images of essential components at the bottom of the moving train.}
        \label{fig4: hardware}
\end{figure}

We adopt six visual fault datasets~\cite{Zhang_TII_21, Zhang_TII_22} from freight trains, including the angle cock, bogie block key, brake shoe key, cut-out cock, dust collector, and fastening bolt on brake beam to assess the effectiveness of our method\footnote{https://github.com/MVME-HBUT/SDD-FTI-FDet}. 
\begin{itemize}
    \item \textbf{Angle Cock} plays a crucial role in the air brake system of freight trains by ensuring the smooth flow of air in the main pipeline. The training and evaluation are performed on 1416/606 images in the train/test in this dataset.
    
    \item \textbf{Bogie Block Key} is placed between the train wheel axle and the bogie to prevent any vibration or displacement that may occur during normal operation. This dataset is split into 5440/2897 images for training/testing.
    
    \item \textbf{Brake Shoe Key} is an important component that is installed in the brake shoe to ensure the safe functioning of the braking system. The dataset provides 3946/1690 images for training/testing.
    
    \item \textbf{Cut-out Cock} plays a vital role in shutting off the airflow between the main fuel tank and the brake pipe. This dataset is divided into 815 and 850 images prepared for training and testing, respectively.
    
    \item \textbf{Dust Collector} is installed on the brake pipe to prevent any abnormal wear or eventual failure of the brake device caused by impurities present in the power source. This dataset contains 583/452 images for training/testing.
    
    \item \textbf{Fastening Bolt (Missing) on Brake Beam} is an important part of revealing the impact load on the brake beam during the braking process. This dataset is divided into 2050/1902 images for training/testing.
\end{itemize}
These six components are critical to ensuring the security of the freight train braking systems. All images are captured by image acquisition equipment installed along and beside the railway, as illustrated in Fig.~\ref{fig4: hardware}.
\subsubsection{Evaluation Metrics}
We consider average precision (AP) with different intersection over union (IoU) thresholds from 0.5 to 0.95 with an interval of 0.05 as the evaluation metric~\cite{metric}, \textit{i.e.}, mAP, AP$_{50}$, AP$_{75}$ to validate the accuracy of the proposed method. Moreover, we also use average recall (AR) that is the maximum recall given a fixed number of detections per image, averaged over categories and IoUs, \textit{i.e.}, AR$_{1}$, AR$_{10}$. We employ multiple metrics to evaluate the efficiency of our framework, such as training cost and inference time. To assess the suitability of the model for resource-limited scenarios, we also keep track of memory utilization and model size.
\subsubsection{Implementation Details}
We first conduct experiments on the ImageNet-1K benchmark~\cite{imagenet} for our FaultMLP under the same training strategy~\cite{AS-MLP}, which uses AdamW optimizer for 300 epochs with a batch size of 1024 and an initial learning rate of 0.001 with cosine decay and 20 epochs of linear warm-up. The pre-trained FaultMLP for ImageNet is used to initialize shared convolutional layers of our backbone network. For six fault datasets, all models are trained for 18K iterations on a single NVIDIA GTX2080ti GPU with an initial learning rate of 0.01.  In order to reduce the instability of the network in the early stage of training, we adopt a warm-up strategy in the first 1000 iterations, gradually increasing the learning rate from $ 10^{-5} $ to $ 10^{-2} $. During training, the inputs are modified to 700 $\times$ 512 pixels, and we apply no data augmentation during testing. For simplicity, we use R18, R50, and R101 to represent ResNet-18, ResNet-50, ResNet-101. We use an SGD optimizer with 0.9 momentum and $10^{-4}$ weight decay. In the ablation experiments, we use a batch size of 4 for training. Additionally, the hyperparameters are fine-tuned during the training phase to achieve the best results about each model in Table~\ref{sota}.

\subsection{Comparison with State-of-the-Art Methods}
\subsubsection{Comparison with General Object Detection}
To further illustrate the superiority, our method is compared with the most popular object detection methods. Notably, the parameter settings of these detectors were all fine-tuned on our device for optimal performance. Table~\ref{sota} shows that most current detectors can hardly achieve the real-time detection speed (30 frames) required by the train fault detection task on the six fault datasets. Notably, CenterNet~\cite{Centernet} and YOLOF~\cite{YOLOF} satisfy real-time detection requirements in speed, but CenterNet still lacks accuracy. The accuracy of YOLOF is almost the best, but the model size is approximately 2.3$\times$ beyond ours.
\subsubsection{Comparison with Teacher-Student KD}
In Table~\ref{sota} and ~\ref{cost}, our method is compared with state-of-the-art teacher-based distillation methods and teacher-free SOTA methods. In comparison to the conventional teacher-based distillation approach, our SDD FTI-FDet supplies the greatest detection accuracy and the smallest model size. The detection accuracy of CWD~\cite{CWD} is comparable to ours, but the model size is almost 5.4$\times$ that of ours. At the same time, the teacher-based method needs to implement the distillation process based on a powerful teacher model, which additionally increases the training cost and reduces the training efficiency. This may bring a heavier training burden in practical applications, and it is not friendly enough for low-resource scenarios such as train fault detection. In terms of speed, RKD~\cite{ReviewKD} is slightly faster than ours, but it cannot meet the needs of the actual scene of train fault detection in terms of model size and detection accuracy. 
Our approach offers two distinct advantages over the latest teacher-free SOTA method. Specifically, it provides an enhanced ability to capture spatial information and utilizes a more efficient MLP architecture, resulting in lightweight and highly effective models.
\subsubsection{Comparison with Teacher-Free KD}
In Table~\ref{sota}, our model size is only 74.5MB, which is 80$\%$ smaller than LabelEnc/LGD while still achieving 4.5$\%$/0.4$\%$ mAP gain. In addition, although LabelEnc has no pre-trained teachers, the two-stage training mode still leads to low training efficiency. For LabelEnc/LGD, our method saves 63$\%$/43$\%$ training time and performs better. According to the results of experiments on six fault datasets, our method performs competitively in terms of accuracy, training expense, and model size and is better suited to train fault detection scenarios.
\subsubsection{Comparison of Training Cost with Other KD Strategies}
As shown in Table~\ref{cost}, the proposed SDD FTI-FDet achieves the best performance, outperforming all KD-based methods. The accuracy of CWD~\cite{CWD} is consistent with our method, but its overall training cost (GPU hours) is more than 3.5$\times$ that of ours. As described in LabelEnc~\cite{LabelEnc}, a self-knowledge distillation technique employs a two-step training approach. In the first step, a label encoding function is obtained, followed by the training of the final detector in the second step. Therefore, the training cost of LabelEnc is over 2.7$\times$ that of our method. In addition, the training cost of LGD~\cite{LGD} is slightly higher than our SDD FTI-FDet in which the dynamic teacher makes mAP produce various performance gains. 
\subsubsection{Qualitative Comparison Results}
\begin{figure}[!t]
    \centering 
        \includegraphics[width=3.3in]{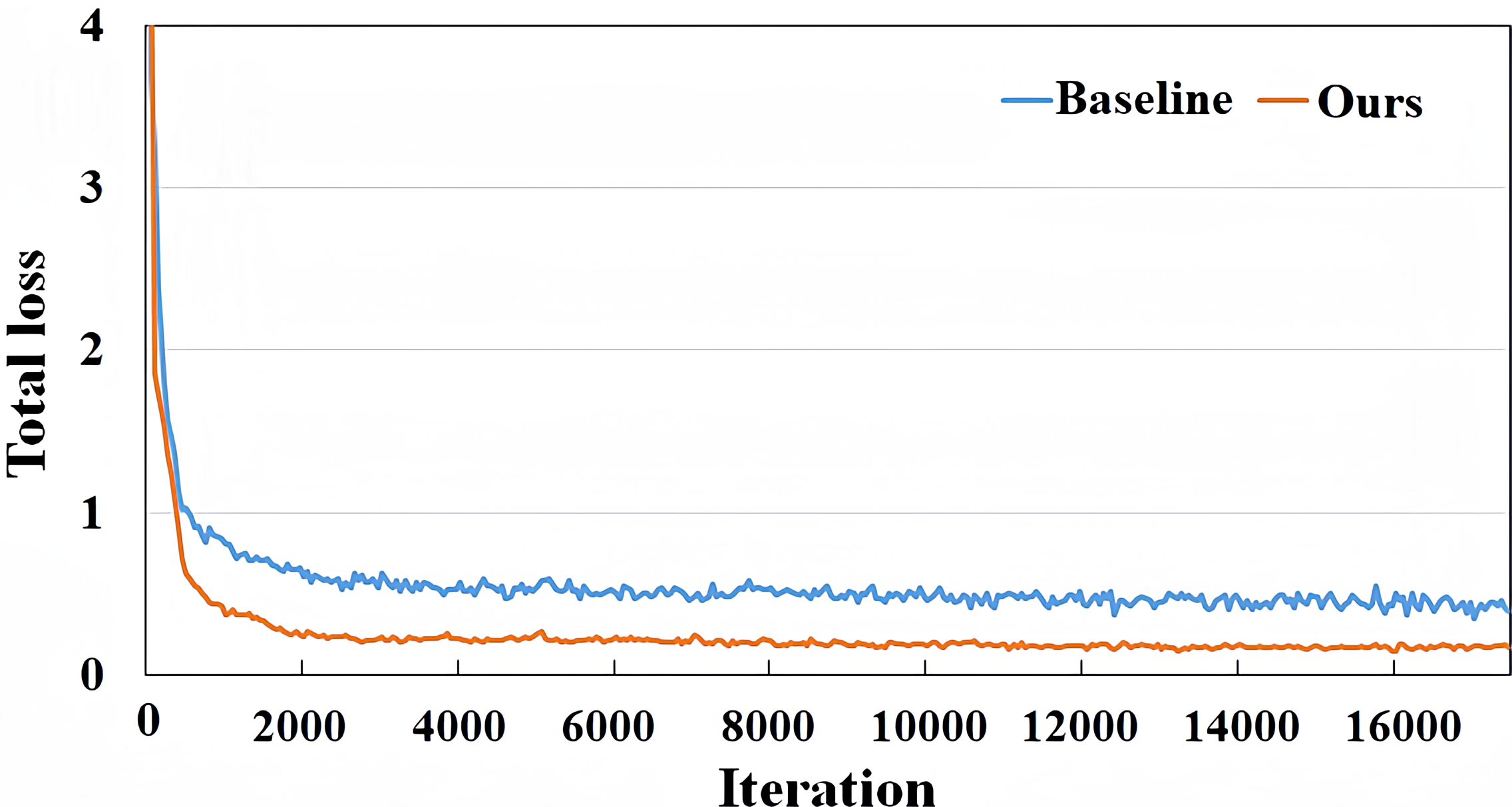}  
	\caption{Performance comparison of total loss between our method and baseline method during training on bogie block key dataset. The total loss curve converges faster indicating that our method is easier to obtain the global optimum. Such phenomenon also illustrates that the proposed dynamic teacher enables joint training of teacher and student, bridging the semantic discrepancy.}
    \label{fig5: curve}
\end{figure}

\begin{figure}[!t]
    \centering    
        \includegraphics[width=3.4in]{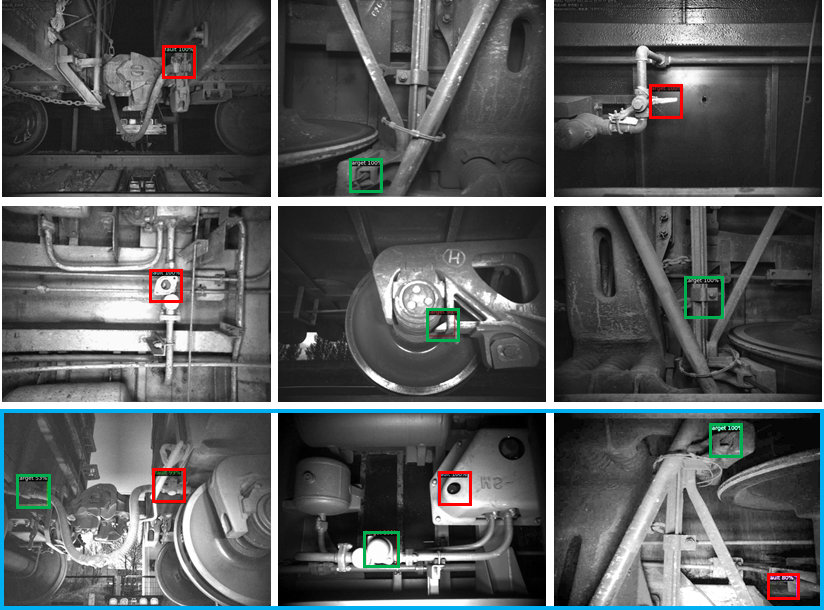}
	\caption{Instance analysis of our method. The \textcolor{green}{green box} is a normal component, while the \textcolor{red}{red box} is a fault one. The top two rows are the correct detection results, and the bottom row is the failure example. The robustness of our method to uneven illumination is not satisfactory. To tackle this issue effectively, we will expand our datasets in the future.
}
    \label{fig: visualization}
\end{figure}

As illustrated in Fig.~\ref{fig5: curve}, our total loss converges faster, making it easier to obtain the global optimum on the fault dataset. The main reason is that the optimized dynamic teacher can enhance the model to bridge the semantic discrepancy while assisting in the supervision of student detectors during training. As a result, the student detector is able to generate comparable predictions with the teacher. Compared with the baseline, our SSD FTI-FDet alleviates the oscillation amplitude of total loss to a certain extent, which verifies the effectiveness of our proposed dynamic teacher. We show the detection results in Fig.~\ref{fig: visualization}, and our approach exhibits limitations in detecting objects accurately when the images are contaminated or have uneven illumination. To enhance the generalization ability, we recognize the crucial role of incorporating these types of images into datasets. Consequently, we will work towards expanding our datasets to alleviate this issue. 
\begin{table}[!t]
    \renewcommand{\arraystretch}{1.0}
    \caption{ Comparison of Training Cost (GPU hours) with other distillation strategies. LabelEnc uses a two-step training pipeline to acquire a label encoding function and then train the final detection model.}
    \centering
    \setlength{\tabcolsep}{1.6mm}{
    \begin{tabular}{lcccccccc}
    \toprule
    \multirow{1}{*}{{Method}}&
    \multirow{1}{*}{{Backbone}}&
    \multirow{1}{*}{{Pre-training}}&
    \multirow{1}{*}{{Finetune}}&
    \multirow{1}{*}{{Overall}}& 
    \multirow{1}{*}{{$mAP$}} \\
    \midrule
    CWD~\cite{CWD}           & R50   & 4.1   & 1.2   & 5.3 & 55.2  \\
    MGD~\cite{MGD}           & R50   & 4.3   & 3.6   & 7.9 & 54.3  \\
    LabelEnc~\cite{LabelEnc} & R50   & 2.8   & 1.3   & 4.1 & 50.7  \\
    LGD~\cite{LGD}           & R50   & --    & 2.6   & 2.6 & 54.8  \\
    \textbf{SDD FTI-FDet}            & \textbf{R50}   & \textbf{--}    & \textbf{1.5}   & \textbf{1.5} & \textbf{55.2}  \\
    \bottomrule
    \end{tabular}}
\label{cost}
\end{table}


\begin{table}[!t]
    \renewcommand{\arraystretch}{1.05}
    \caption{Comparison results with various backbones between CNN and MLP architecture on the bogie block key dataset.}
    \label{backbone}
    \centering
    \setlength{\tabcolsep}{1.5mm}{
    \begin{tabular}{lccccccc}
        \toprule
        \multirow{2}{*}{{Backbones}}&
        \multirow{2}{*}{{$mAP$}}& 
        \multirow{2}{*}{{$AP_{50}$}}& 
        \multirow{2}{*}{{$AP_{75}$}}&
        \multirow{2}{*}{{$AR_{1}$}}&
        \multirow{2}{*}{{$AR_{10}$}}& 
        \multirow{2}{*}{\makecell[c]{Model\\Size (MB)}}
        \\
        & &  & & &  \\
        \midrule
            MobileNetv3~\cite{MobilenetV3} & 46.9  & 95.5 & 39.4 & 55.7 & 56.6 & 229.8 \\
            ShuffleNetv2~\cite{ShufflenetV2}   & 45.4  & 94.5 & 37.0 & 53.4 & 54.6 & 223.8 \\
            R18~\cite{ResNet}    & 48.8  & 97.4 & 41.4 & 56.8 & 58.2 & 305.1 \\
            R34~\cite{ResNet}    & 49.6  & 97.0 & 42.6 & 56.7 & 58.1 & 385.7 \\
            R50~\cite{ResNet}    & 51.8  & 96.5 & 47.2 & 59.4 & 59.8 & 409.5 \\
            SwinT-t~\cite{swint}   & 50.8  & 96.8      & 44.4  & 57.9  & 58.9  & 387.5  \\
            SwinT-s~\cite{swint}   & 49.2  & 97.3      & 42.7  & 57.1  & 58.4  & 608.7  \\
            SwinT-b~\cite{swint}   & 47.2  & 96.6      & 37.3  & 55.7  & 56.8  & 608.9  \\
            ASMLP\_t~\cite{AS-MLP} & 50.5  & 97.7      & 43.8  & 57.7  & 58.5  & 387.3  \\
            ASMLP\_s~\cite{AS-MLP} & 51.4  & 97.9      & 45.3  & 58.3  & 59.2  & 558.0  \\
            ASMLP\_b~\cite{AS-MLP} & 50.7  & 97.8      & 44.5  & 57.7  & 58.7  & 862.2  \\
            \textbf{FaultMLP}(ours)  & \textbf{51.0} & \textbf{97.8} & \textbf{45.1} & \textbf{58.0}   & \textbf{59.0}    & \textbf{238.8}  \\
        \bottomrule
    \end{tabular}}
\end{table}

\begin{table}[!t]
  \renewcommand{\arraystretch}{1.0}
  \centering
  \caption{Performance of MLP-based FPN modules with various channels on the bogie block key dataset. }
    \label{tab5}
    \setlength{\tabcolsep}{2mm}{
    \begin{tabular}{cccccccc}
    \toprule
        \multirow{2}{*}{{Channels}}& 
        \multirow{2}{*}{{$mAP$}}& 
        \multirow{2}{*}{{$AP_{50}$}}& 
        \multirow{2}{*}{{$AP_{75}$}}&
        \multirow{2}{*}{{$AR_{1}$}}&
        \multirow{2}{*}{{$AR_{10}$}}& 
        \multirow{2}{*}{\makecell[c]{Model\\Size (MB)}}  \\
           & & & & & &   \\
    \midrule
    256   & 51.0  & 97.8  & 45.1  &  58.0     &  59.0       & 238.8  \\
    128   & 51.1  & 97.6  & 44.2  &  57.9     & 58.9        & 177.6  \\
    \textbf{64}    & \textbf{51.1}  & \textbf{97.3}  & \textbf{43.6}  &   \textbf{58.0}    & \textbf{59.1}        & \textbf{148.9}  \\
    32    & 50.8  & 97.7  & 44.1  &  58.1     & 59.2        & 135.1  \\
    16    & 49.2  & 97.5  & 42.3  &  57.0     &  58.2       & 128.3  \\
    8     & 50.5  & 97.7  & 44.0  & 58.3  & 59.5    & 124.9  \\
    \bottomrule
    \end{tabular}}
\end{table}

\begin{table}[!t]
  \centering
  \renewcommand{\arraystretch}{1.0}
  \caption{Ablation study of key components of our method. Dim: Various segment dimensions in the feature adaptor. $\tau$: Various temperatures in the dynamic distillation stage. /: Without feature adaptor or dynamic distillation. --: With self-distillation in baseline method. }
  \setlength{\tabcolsep}{2mm}{
    \begin{tabular}{lcccccc}
    \toprule
        \multirow{2}{*}{{Backbone}}& 
        \multirow{2}{*}{{Dim}}& 
        \multirow{2}{*}{{$\tau$}}& 
        \multirow{2}{*}{{$mAP$}}& 
        \multirow{2}{*}{{$AP_{50}$}}&
        \multirow{2}{*}{{$AR_{1}$}}&
        \multirow{2}{*}{{$AR_{10}$}}  \\
          & & & & & &    \\
    \midrule
    \multirow{2}{*}{\makecell[c]{R50\\(Baseline)}} &   \multirow{2}{*}{/}    & \multirow{2}{*}{--}     & \multirow{2}{*}{51.8}  & \multirow{2}{*}{96.5}  & \multirow{2}{*}{59.4}  & \multirow{2}{*}{59.8} \\ 
    & & & & & &   \\
    \midrule
    \multirow{10}[10]{*}{\textbf{FaultMLP}} &  /     &  --     & 51.1  & 97.3  & 58.0      & 59.1     \\
         & 1 & --      & 50.4  & 97.7  & 57.6  & 59.0   \\
         & 2 &   --    & 51.0  & 97.5  & 58.5  & 59.6    \\
\cmidrule{2-7}          & \multirow{6}[2]{*}{\textbf{4}} 
                  & / & 51.3  & 97.7  & 58.2  & 59.3 \\
          &       & --  & 51.5  & 97.7  & 58.6  & 59.8 \\
          &       & 1   & 51.0  & 97.7  & 58.5  & 59.5   \\
          &       & 5   & 51.5  & 97.6  & 58.4  & 59.8   \\
          &       & 10  & 51.3  & 97.1  & 58.2  & 59.4   \\
          &       & \textbf{15} & \textbf{52.3} & \textbf{98.4} & \textbf{59.2} & \textbf{60.5}  \\
          &       & 20  & 51.5  & 97.7  & 58.4  & 59.6   \\
\cmidrule{2-7}          & 8 &   --    & 51.1  & 97.3  & 58.2  & 59.3  \\
    \bottomrule
    \end{tabular}}
    \label{tab6}
\end{table}

\subsection{Ablation Study}\label{sec:ablation}
\subsubsection{MLP-based Backbone}
We experimented with various backbones to explore a more efficient framework to replace the baseline networks. To meet the needs of low-resource scenarios, we use precision, recall, training time, inference time, and model size as evaluation metrics for selecting an appropriate backbone. In  Table~\ref{backbone}, ASMLP-s provides the best accuracy in mAP (0.4$\%$ higher than FaultMLP), but its model size is 2.3$\times$ that of FaultMLP, which is unfriendly to low-resource scenarios. SwinT-t needs greater training time and memory even if its accuracy and model size are similar to those of FaultMLP. Therefore, FaultMLP more effectively satisfies the requirements of low-resource situations. 
\subsubsection{Channels in MLP-based FPN}
Since most of the key components in train fault images are relatively small in size, we adjust the number of channels in FPN to reduce redundant information in the model. In Table~\ref{tab5}, the accuracy of the model rises when the channel is decreased from 256 to 128/64, demonstrating the success of our attempt. At the same time, when the number of channels is 64, the model size is reduced by 30MB compared to 128, which is more in line with the requirements of field deployment. The above data shows that 64 channels are the optimal choice for fault detection.
\subsubsection{Dimensions in Feature Adaptor}
As described before, our method adopts encoding in three directions for projection and then stitching to obtain features with more spatial information. As shown in Table~\ref{tab6}, we first try to discretize the spatial information into finer-grained features, and the network achieved an accuracy of 51.1$\%$ when $dim=8$. However, redundant information cannot be avoided in the process of projection aggregation. Therefore, we evaluated how well the network performed utilizing various dims on the bogie block key dataset. We observe that when $dim=4$, the features learned by the network are more accurate, comprehensive, and less redundant.
\subsubsection{Temperature $\tau$ in Distillation Loss}
We utilize $\lambda$ in Eq.~(\ref{Eq1}) to balance the distillation and hard label loss, among which the hyperparameter temperature $\tau$ to tune the probability distribution of the dynamic teacher output. Table~\ref{tab6} reports the results of distillation at different temperatures. Compared to the first row, we find that students perform better when the temperature $\tau$ becomes higher. But when $\tau$ is too large, the probability distribution of the dynamic teacher output tends to be flattened, and the student model is more susceptible to the noise in the negative labels, affecting performance improvement. In this paper, we simply set the $\tau$ to 15, which is fixed in all experiments.

\subsubsection{Framework Analysis}
To explore our final structure, we first investigate the performance of different modules. In Table~\ref{backbone}, we find that the ResNet50 performs better than ours. This indicates that ResNet50 can capture more useful information, but the model size is extremely larger. For the role of dynamic distillation strategy, the results demonstrate that conducting feature adaptor and its with dynamic distillation loss can improve the accuracy of the model by +0.2 and +1.2 mAP, respectively. This indicates that the proposed dynamic distillation strategy in our framework is indeed effective. The main reason is that the MLP-based feature adaptor is helpful in modeling the intricate spatial information between channels for the dynamic teacher. Subsequently, dynamic (logit) distillation causes to enforce the student network outputs more similar predictions with the teacher network.

\section{Conclusion}\label{sec: conclusion}
This paper introduces a highly effective framework called SDD FTI-FDet, which incorporates an axial shift strategy that enables MLP-like architectures to circumvent the limitation of space invariance on fault detection. Furthermore, the proposed dynamic distillation method adopts an efficient encoding mechanism to model abundant spatial and semantic information without relying on pre-trained teachers. Extensive experiments on six visual fault datasets from freight trains show that the joint training between dynamic teacher and student networks can significantly decrease the training cost and improve approximately 5.2$\times$ for the training efficiency. Meanwhile, our method delivers the highest accuracy real-time detection (38.8 FPS) while meeting the low computational resources compared with the state-of-the-art methods.

In the future, we intend to focus on strengthening our proposed framework to accomplish real-time detection for multi-fault on different embedded devices (\textit{e.g.} Raspberry Pi and Jetson Nano).

\bibliographystyle{IEEEtranTIE}
\bibliography{IEEEabrv,BIB_xx-TIE-xxxx}

\vfill

\end{document}